\documentclass[sigconf, natbib=true, anonymous=false]{acmart}
\usepackage{graphicx}
\usepackage{subfigure}
\usepackage{multirow}
\usepackage{ulem}
\usepackage{bbding}
\usepackage{makecell}
\AtBeginDocument{%
  \providecommand\BibTeX{{%
    \normalfont B\kern-0.5em{\scshape i\kern-0.25em b}\kern-0.8em\TeX}}}

\setcopyright{acmcopyright}
\copyrightyear{2022}
\acmYear{2022}
\acmDOI{XXXXXXX.XXXXXXX}

\acmConference[Conference acronym 'XX]{Make sure to enter the correct
  conference title from your rights confirmation emai}{June 03--05,
  2018}{Woodstock, NY}
\acmPrice{15.00}
\acmISBN{978-1-4503-XXXX-X/18/06}

\begin{document}

\title{Generation-Guided Multi-Level Unified Network for Video Grounding}

\author{Xing Cheng}
\affiliation{%
  \institution{MMU KwaiShou Inc.}
  \city{Beijing}
  \country{China}
}
\email{chengxing03@kuaishou.com}

\author{Xiangyu Wu}
\affiliation{%
  \institution{MMU KwaiShou Inc.}
  \city{Beijing}
  \country{China}
}
\email{wuxiangyu@kuaishou.com}

\author{Dong Shen}
\affiliation{%
  \institution{MMU KwaiShou Inc.}
  \city{Beijing}
  \country{China}
}
\email{shendong@kuaishou.com}

\author{Hezheng Lin}
\affiliation{%
  \institution{MMU KwaiShou Inc.}
  \city{Beijing}
  \country{China}
}
\email{linhezheng@kuaishou.com}

\author{Fan Yang}
\affiliation{%
  \institution{MMU KwaiShou Inc.}
  \city{Beijing}
  \country{China}
}
\email{yangfan@kuaishou.com}

\renewcommand{\shortauthors}{Trovato and Tobin, et al.}

\begin{abstract}
Video grounding aims to locate the timestamps best matching the query description within an untrimmed video. Prevalent methods can be divided into moment-level and clip-level frameworks. Moment-level approaches directly predict the probability of each transient moment to be the boundary in a global perspective, and they usually perform better in coarse grounding. On the other hand, clip-level ones aggregate the moments in different time windows into proposals and then deduce the most similar one, leading to its advantage in fine-grained grounding. 

In this paper, we propose a multi-level unified framework to enhance performance by leveraging the merits of both moment-level and clip-level methods. Moreover, a novel generation-guided paradigm in both levels is adopted. It introduces a multi-modal generator to produce the implicit boundary feature and clip feature, later regarded as queries to calculate the boundary scores by a discriminator. The generation-guided solution enhances video grounding from a two-unique-modals' match task to a cross-modal attention task, which steps out of the previous framework and obtains notable gains. The proposed Generation-guided Multi-level Unified network (GMU) surpasses previous methods and reaches State-Of-The-Art on various benchmarks with disparate features, e.g., Charades-STA, ActivityNet captions.
\end{abstract}

\ccsdesc[500]{Information system~Novelty in information retrieval; Multimedia and multimodal retrieval}

\keywords{Video grounding; cross-modal; generation; multi-level;}


\maketitle



\section{Introduction}

\begin{figure}[ht]
\begin{center}
\includegraphics[scale=0.40]{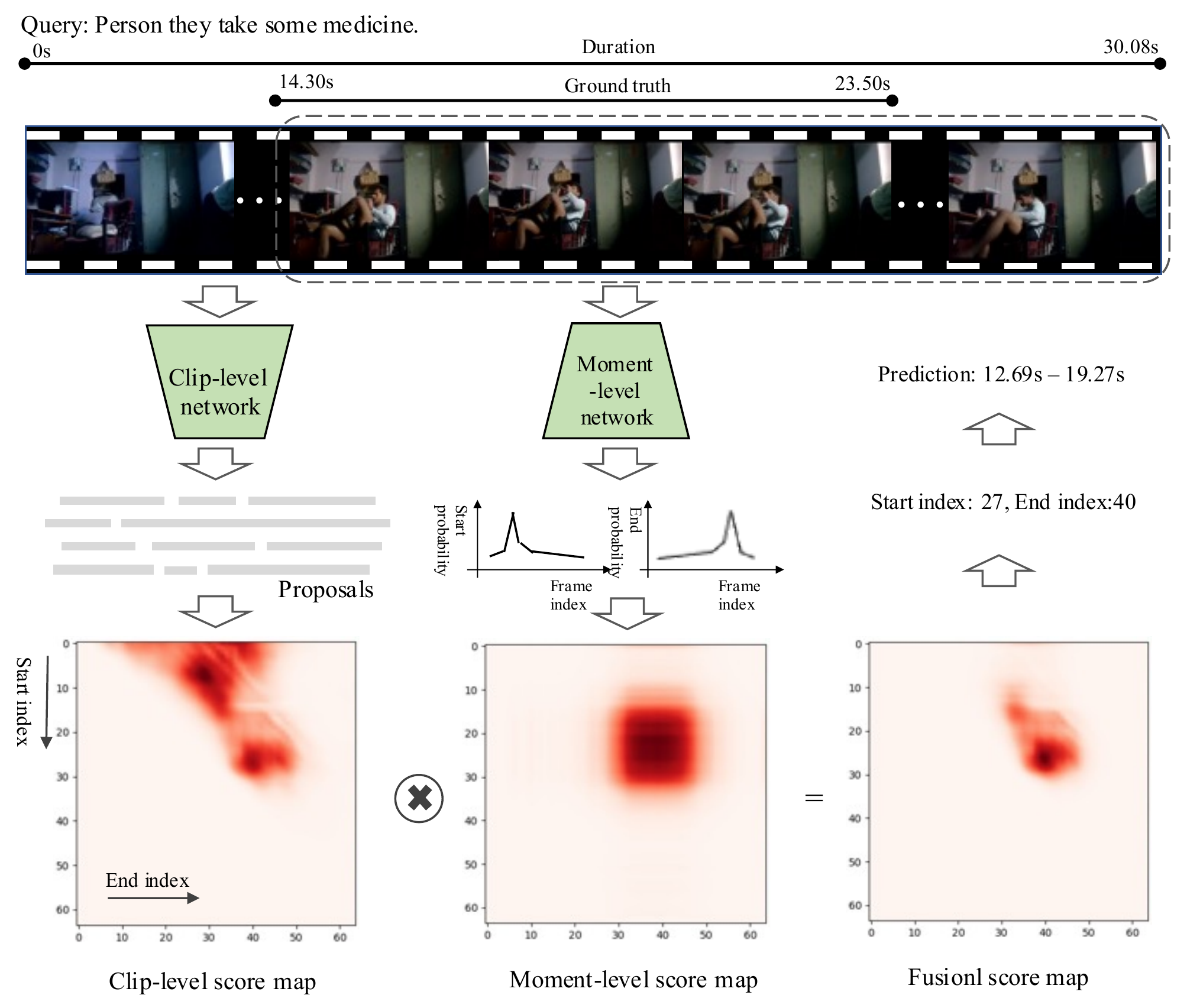}
\end{center}
\caption{The introduction of the video grounding task and the overall framework of the proposed GMU. The redder points represent the higher the predicted score. The top-1 score in the clip-level map falls into the wrong cluster, and the moment-level map is blurry in deducing accurate results. A unification of them solves both defects.}
\label{framework}
\end{figure}


Video grounding \cite{gao2017tall}, as one of the cross-modal tasks \cite{zhang1997integrated, luo2021clip4clip, gao2017video, pan2017video, karpathy2014large, yue2015beyond}, aims to precisely locate the timestamps corresponding to given queries, e.g., "Person they take some medicine." as shown in Fig. \ref{framework}. We need to deduce the accurate timestamp [14.3, 23.5] denoting the motion "take some medicine". Previous works can be roughly categorized as moment-level and clip-level. 

\begin{figure}[h]
\begin{center}
\subfigure[A diagram of previous moment-level framework.]{\includegraphics[scale=0.28]{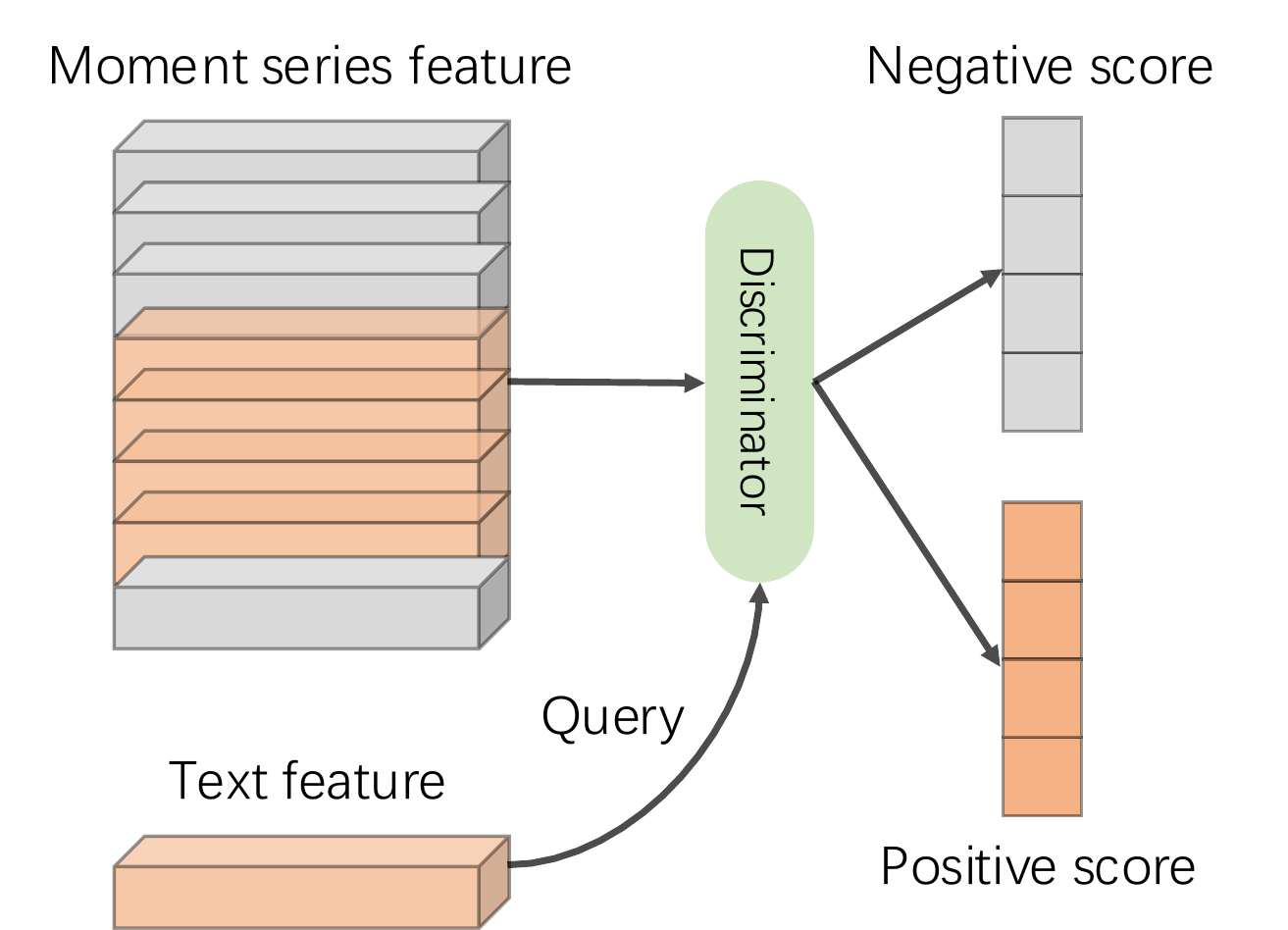}\label{frame_ori}}
\subfigure[A diagram of previous clip-level framework.]{\includegraphics[scale=0.27]{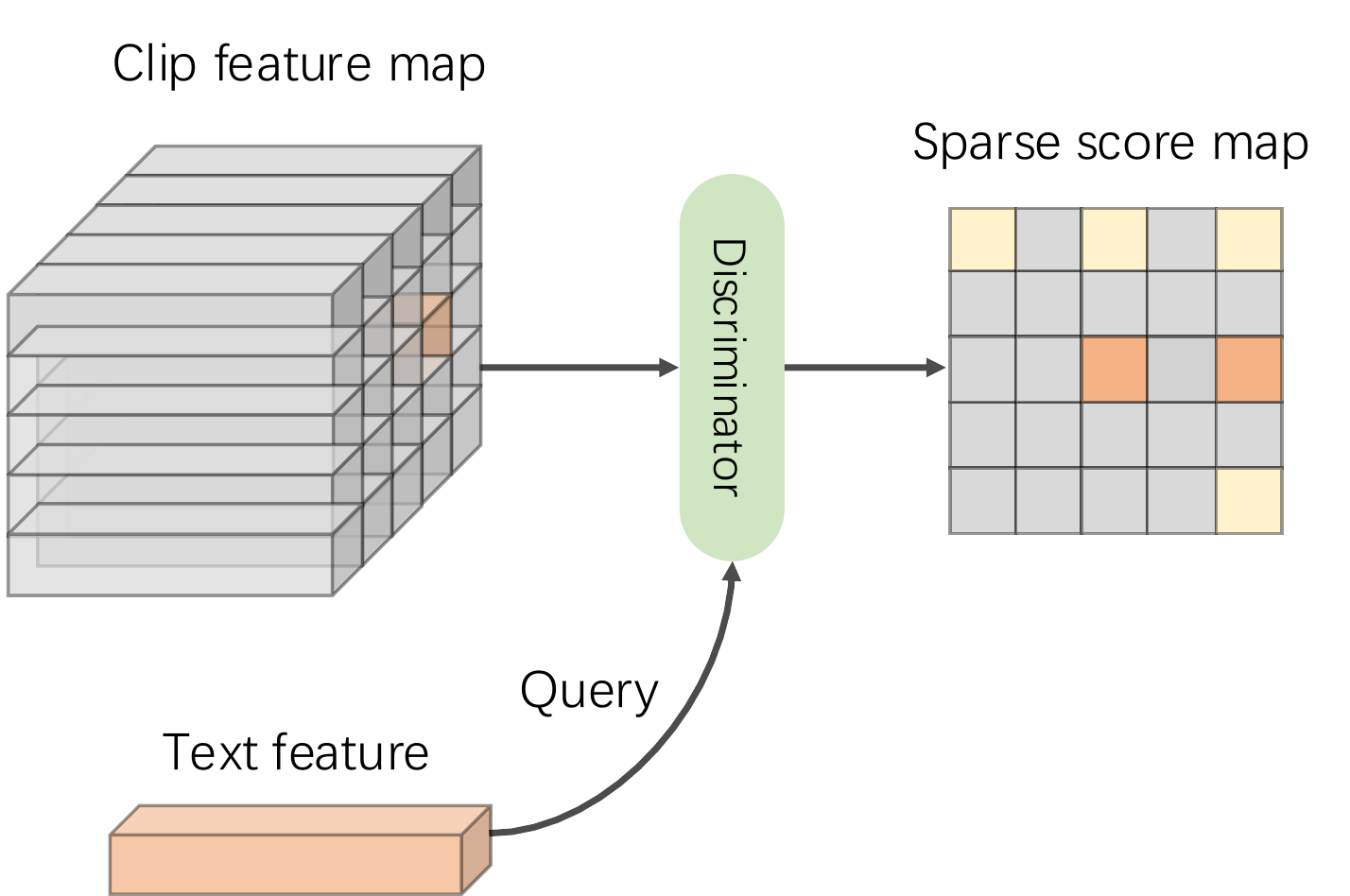}\label{clip_ori}}
\subfigure[A diagram of the generation-guided moment-level framework.]{\includegraphics[scale=0.21]{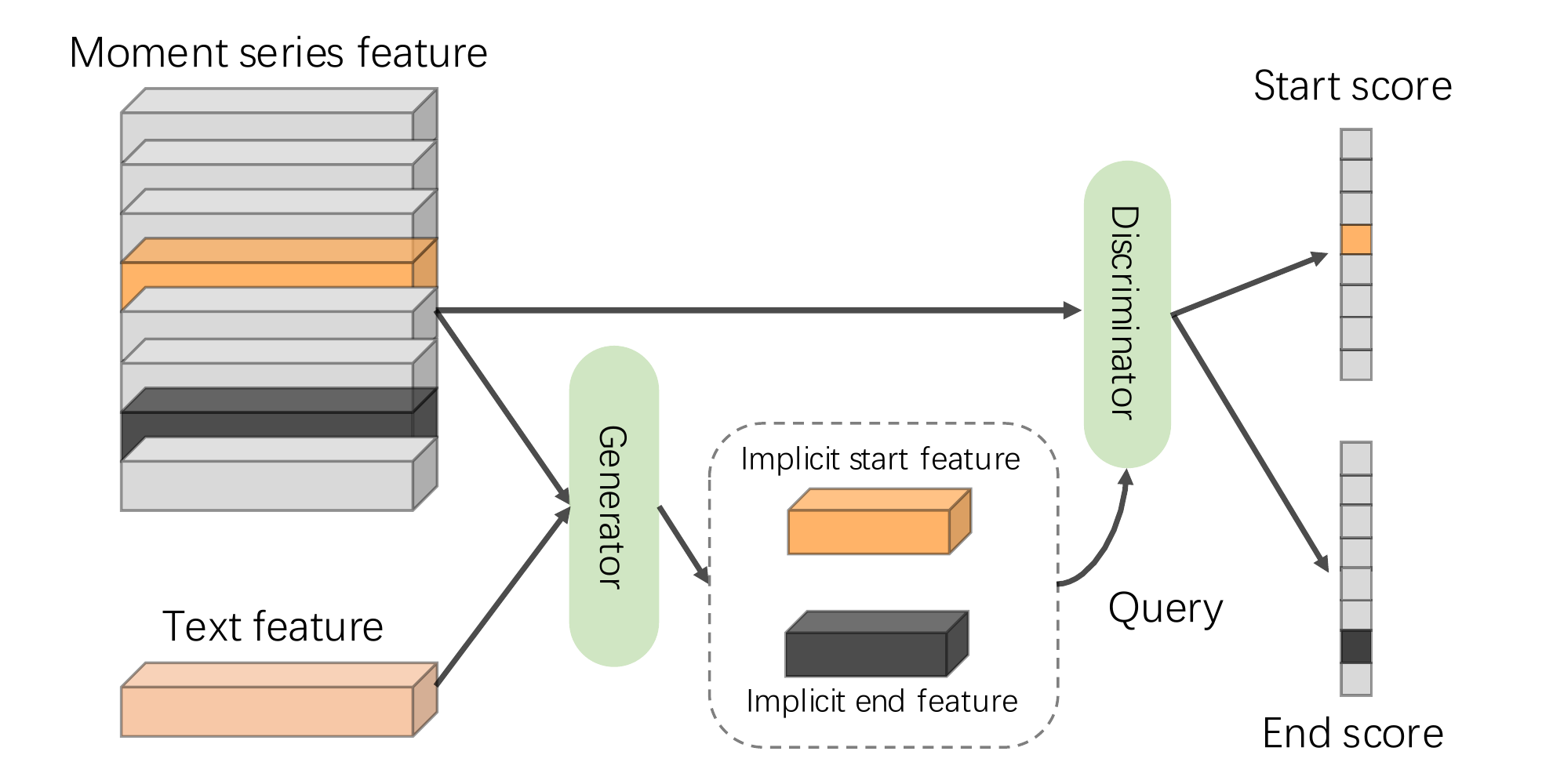}\label{frame_ours}}
\subfigure[A diagram of the generation-guided clip-level framework.]{\includegraphics[scale=0.21]{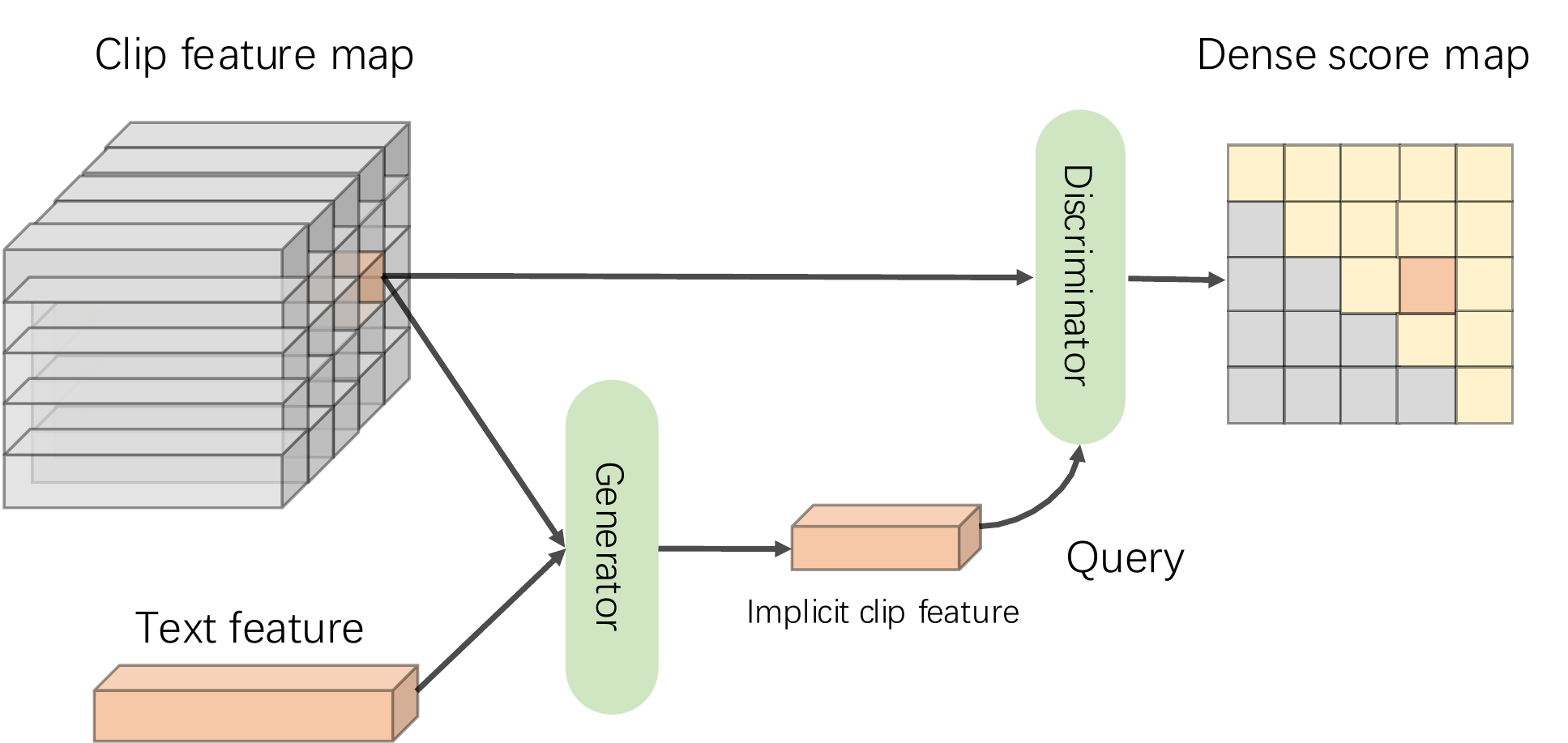}\label{clip_ours}}
\end{center}
\caption{The comparisons of the proposed generation-guided frameworks with previous straightforward methods. In the maps of clip-level results, each point denotes a proposal, the map's horizontal axis represents the start index and the vertical axis represents the end index. The yellow points are the proposals taken into consideration, and the orange points are the final results which model chooses from yellow points.}
\end{figure}

Moment-level methods \cite{mun2020local} process the whole video in a global perspective and then deduce the localization or judge the probability of each moment to be the boundary as shown in Fig. \ref{frame_ori}. A biaffine multiplication \cite{2016Deep} of the start and end probability leads to the score map, which performs well in coarse grounding but is poor in fine-grained grounding due to the separate predicting strategy. Clip-level methods \cite{zhang2020learning} segment the video into diverse clips and then retrieve the most similar proposal, which is called clip-level. It immediately simulates the Ground Truth(GT) Intersection-over-Union(IoU) distribution and generates several clusters around the diagonal because the overall GT exists around the diagonal as explored in \cite{2021}, a wrong cluster then leads to great deviation referring to Fig. \ref{framework}, which is weaker than moment-level at coarse grounding while on the opposite at fine-grained grounding. We conduct extensive experiments in Sec. \ref{result_compare} and visualization in Sec. \ref{visualization} to support the standpoint.

In this paper, we introduce a multi-level feature fusion mechanism to leverage the merits of both methods. As shown in Fig. \ref{framework}, we first sample the frames within the video at a certain sampling rate. Pre-trained C3D \cite{tran2015learning}, I3D \cite{carreira2017quo}, or VGG \cite{simonyan2014very} are employed to extract their features. Clip-level components enumerate all possible proposals and measure their scores. Moment-level components directly predict the probability of each moment being the boundary. Finally, these score maps are fused together using the Hadamard product, producing a more convincing result. Further experiments prove that such an improved result is not directly caused by the fusion step itself but rather a benefit from the enhanced visual and semantic representation of the model by incorporating the unified training steps.

Moreover, we adopt a generation-guided architecture as shown in Fig. \ref{frame_ours} and Fig. \ref{clip_ours}. In previous approaches of either clip-level or moment-level, the video grounding task is processed in visual modal, and text modal independently in the feature encoding phase, and then they conduct a match strategy upon the encoded features \cite{mun2020local, zhang2020learning}. Learning a consistent latent space can be burdensome because of the heterogeneity between different modals. We devise generators to produce the target feature in both levels, where target means boundary at the moment level and target clip at the clip level. The target features are cross-modal representations and share the unified latent space with visual features. Later, they are fed into the discriminator as queries, while the original visual features are values to be matched. Such architecture significantly alleviates the cross -modal heterogeneity, and the ablation experiments show it is effective in video grounding tasks.

Last, previous clip-level methods commonly adopt the sparse predicting strategy. For example, as shown in Fig. \ref{clip_ori}, the ground truth is the orange point in the clip feature map with a coordinate of $(2,3)$. Because only yellow points are taken into consideration with sparse predicting, it will make a prediction of $(2,2)$ or $(2,4)$, indicating a loss of accuracy. To solve that, we sample frames sparsely with sampling mask weight in BMN \cite{lin2019bmn} and adopt a dense predicting strategy shown in Fig. \ref{clip_ours}, where all points in the score map are contributed to the loss function. This strategy brings about a significant enhancement of accuracy. Our contributions can be concluded as follows:
\begin{itemize}
    \item We propose a multi-level unified framework that fuses both moment-level and clip-level information and improves the representation of the based module.
    \item Casting off the normal practice, we adopt generation-guided architecture in both levels to enhance the alignment of cross-modal feature, and introduce an attention mechanism to optimize the video grounding task.
    \item With multi-level unification, generation-guided, and some verified tricks like Mask Language Model (MLM) and dense predicting, the proposed Generation-guided Multi-level Unified framework (GMU) outperforms the baselines in both moment-level and clip-level components and reaches the State-Of-The-Art on Charades-STA, ActivityNet caption.
\end{itemize}

\section{Related work}

In this section, we first introduce previous moment-level \cite{ mun2020local, zeng2020dense, nan2021interventional, zhang2021video, ghosh2019excl, lu2019debug, wang2020temporally} and clip-level \cite{zhang2020learning, zhang2021multi, gao2021fast, zhang2019cross, yuan2019semantic, zhang2019man} methods in detail. Then some fusion methods considering both levels and generation methods in some other cross-modal fields are also presented.


\textbf{Moment-level.} It is initially proposed in \cite{gao2017tall} sliding windows with fixed length are adopted to segment videos into short clips, which are sent to the feature extractor, followed by multi-modal intersection module and a regressor for location inference. LGI \cite{mun2020local} assumes that contextual information is hidden in mid-level semantic phases, which consist of several sequential words. It designs diverse local attention mechanisms for multi-level phases, and they are consequently unified by global context intersection. DRN \cite{zeng2020dense} stacks the feature pyramid to multiply and intersect the visual and semantic representations. Each layer contains a grounding module with three heads calculating the matching score, IoU score, and location score, respectively.

\textbf{Clip-level.} 2D-TAN \cite{zhang2020learning} enumerates all timestamps and aggregates the frame embedding within them into fixed-dimension features with pooling or convolutional operation. Then a 2D temporal map is formed, where every timestamp's representation can be found. The original location problem is transformed into a retrieval task. Subsequently, MS-2D-TAN \cite{zhang2021multi} and FMVC \cite{gao2021fast} based on 2D-TAN are proposed. MS-2D-TAN expands its single-scale into multi-scale encoding. FMVC employs hierarchical semantic-guided attention to exploit the local-global dependency upon the semantic role tree, which can be better matched with proposals.

\textbf{Fusion-level.} DPIN \cite{10.1145/3394171.3413975} is the first to propose an multi-level integration framework. BPNet \cite{xiao2021boundary} adopts the same strategy with cross-modal attention. LPNet \cite{xiao2021natural} devises a set of hand-designed moment candidates in advance to exploit the prior knowledge. They all achieve a huge improvement in effectiveness.

\textbf{Generation methods.}
Cross-modal generation tasks involving vision and natural language include image captioning \cite{you2016image, farhadi2010every}, video captioning \cite{donahue2015long,gao2017video}, and text-to-image generation \cite{mansimov2015generating,reed2016generative}. They commonly adopt an encoder to transfer the input into an embedding and then explicitly decode it into another modal. In video grounding, there is no immediate need for the generation. This paper, however, focuses on the generation framework, where we replace the original decoder with a discriminator and utilize the generated embedding to implicitly retrieve the optimal match. This framework realizes a significant boost compared to previous architectures.

\section{Method}
\begin{figure*}[ht]
\begin{center}
\includegraphics[scale=0.29]{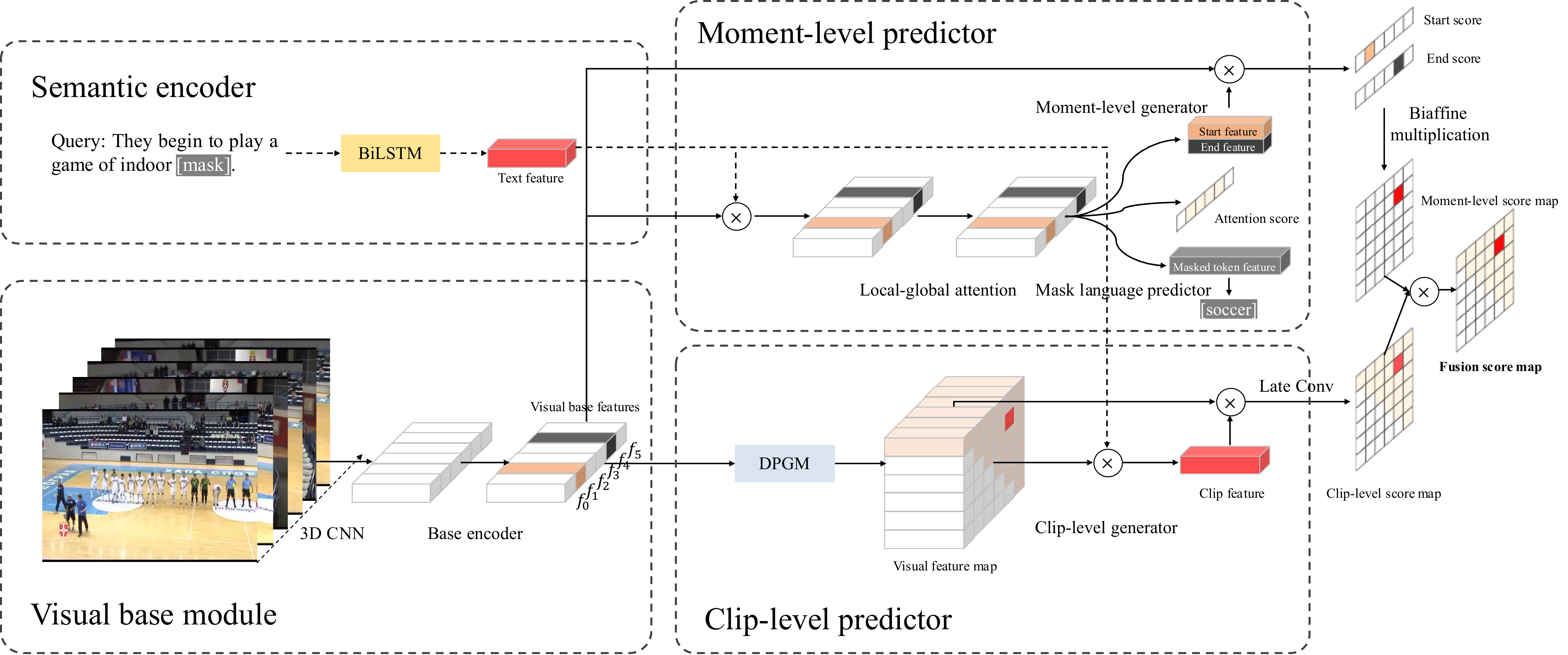}
\end{center}
\caption{
The architecture of the proposed GMU. The text is randomly masked and then encoded by Long Short-Term Memory (LSTM). Visual frames are encoded by pre-trained 3D CNN following the introduced based Feedforward Neural Networks (FNN). For the moment-level, local-global attention and a generator are devised to produce implicit start, end feature, attention score, and masked token feature. The similarity between the start, and end features and short-cut visual base features are followed by FNN to output the start, and end scores, whose biaffine multiplication generates the moment-level score map. For the CLP, a dense proposal generation module (DPGM) is employed to enumerate all possible clips comprised of successive frames. After the fusion of text features and the visual feature map, the fused features are further encoded by clip-level generator and produce an implicit clip feature, which is the query to retrieve the most similar proposal in the short-cut visual feature map. The Hadamard product of moment-level and clip-level score map is the inference output of GMU.}
\label{GMU}
\end{figure*}

The overview of GMU is shown in Fig. \ref{GMU}. It contains four components: semantic encoder, visual base module, moment-level predictor, and clip-level predictor. They are utilized for encoding the text, encoding the visual frames, obtaining the moment-level score map, and obtaining the clip-level score map separately. We will introduce each module in detail in this section with LGI as the moment-level baseline and 2D-TAN as the clip-level baseline.

\subsection{Semantic Encoder}
For fair comparisons with previous methods \cite{mun2020local}, we adopt Bidirectional LSTM (BiLSTM) as the semantic encoder. For the input Query $Q = \{q_1, ..., q_L\}$ with L words, BiLSTM encodes it into semantic feature $s$. To ensure that the top network retains the semantic information, we supplement with the Masked Language Model (MLM) task. We randomly mask one word $q_i$ into $[mask]$, representing the masked token, which is the target to be predicted at the output side.
BiLSTM outputs the sentence feature with its merits to adaptively exploit the long short-term dependency: $s = BiLSTM(Q)$
, where $s \in \mathbb{R}^d$.

\subsection{Visual Base Module}
This component encodes the videos into shared visual base features $V_f$ for both the moment-level and clip level predictor. For the input video $V$, it is first segmented into moments $\{v_1, ..., v_T\}$ with a fixed interval, where $T$ denotes the number of moments. They are extracted into $V_{o} \in \mathbb{R}^{T \times d_i}$ by pre-trained 3D Convolutional Neural Networks (CNN): $V_{o} = 3D\_CNN(V)$, where $d_i$ is the initial dimension. Then they are encoded into $V_f = \{f_1, ..., f_T\}$ by the devised based encoder. Different from 2D-TAN \cite{zhang2020learning}, whose base encoder consists of only FNN, our work owns additional learnable position embedding. This process can be expressed as:

\begin{eqnarray}
V_f = ReLU(V_{o}W_{base} + W_{pos}) 
\end{eqnarray}

Where $W_{base} \in \mathbb{R}^{d_i \times d}$ denotes the parameters in the linear layer, $W_{pos} \in \mathbb{R}^{T \times d}$ denotes the position embedding.

\subsection{Moment-level Predictor}
In this component, the input $V_f$ is first fused with semantic feature $s$ by Hadamard product to obtain the semantic-aware features $f_{sa}$. Then it is fed into local and global attention networks successively. Thus we produce the overall features $f_o$ with both a global perspective and semantic information. An elaborate moment-level generator is then devised to aggregate them into implicit start feature $f_{start}$, implicit end feature $f_{end}$, attention score $s_{att}$, and masked token feature $s_{mask}$. The similarity between the generated start $f_{start}$, end features $f_{end}$, and $V_f$ can spontaneously lead to the start scores $s_{start}$, end scores $s_{end}$. The overall process can be expressed as:
\begin{eqnarray}
&f_{sa} = V_f \odot s\\
&f_o = Attention(f_{sa}) \\
&f_{start},\ f_{end},\ s_{att},\ s_{mask}= Gen_{fl}(f_o) \\
&s_{start} = Sim(V_f,f_{start})\\
&s_{end} = Sim(V_f,f_{end})\\
&S_m = BFM(s_{start}, s_{end})
\end{eqnarray}
Where $Attention$, $Gen_{fl}$ denote the local-global attention and moment-level generator, which are introduced below. $BFM$ denotes the Biaffine Multiplication, which means $S_m^{i,j} = \sqrt{s_{start}^i, s_{end}^j}$. $Sim$ denotes the function measuring the similarity. It can be expressed as:
\begin{eqnarray}
Sim(a,b) = W_a(||a|| \odot ||b||)W_b
\end{eqnarray}
Where $a\in \mathbb{R}^{d}, b \in \mathbb{R}^{T\times d}$ denote the input. $W_a \in \mathbb{R}^{1\times T}$ and $W_b \in \mathbb{R}^{d\times d}$ denote the learned parameters of fully connected layers.

\subsubsection{Local-global Attention}

The semantic-aware feature is encoded by ResBlock as the local attention: $f_{l} = ResBlock(f_{sa})$, 
where $f_{l}$ denotes the output of local attention. $ResBlock$ denotes $n_l$-layer basic block with residual connection, which can be expressed as:
\begin{eqnarray}
&f_{l}^{i} = ReLU(f_{l}^{i-1} + BasicBlock(f_{l}^{i-1}))
\end{eqnarray}
where $f_{l}^{i}$ denotes the $i$-th layer output in the ResBlock. $BasicBlock$ denotes the convolutional networks following the Batch Normalization (BN) and ReLU. The kernel size, stride, and padding of the convolutional networks are set to \{1, 1, 0\}.

We utilize $n_g$-layer None-Local Block (NLBlock) to further exploit the global dependency:
\begin{align}
f_{o}^{i} &= NLBlock(f_{o}^{i-1}) \notag \\
&= f_{o}^{i-1} + (W_{vl}f_{o}^{i-1})Softmax(\frac{(W_{qr}f_o^{i-1})^T(W_{k}f_o^{i-1})^T}{\sqrt{d}})^{T}
\end{align}
where $W_{vl}$, $W_{qr}$, and $W_{k}$ are learnable matrices representing the value, query, and key linear layer. $f_o^i$ denotes the global feature of the output of $i$-th NLBlock. In this case, $f_{o}^{0} = f_{l}^{n_l}$, the input of the global attention equals the output of the local attention.

\subsubsection{Moment-level generator}
In this component, we adopt an Attentive Pooling strategy as \cite{mun2020local} to aggregate the multi-frame features $f_o \in \mathbb{R}^{T \times d}$ into one feature $f_a \in \mathbb{R}^{d}$, and then project it into $f_{start}$, $f_{end}$, and $s_{mask}$ by FNN.
\begin{eqnarray}
&w_a = Softmax(FNN_{AP}(f_o))\\
&f_a = w_af_o\\
&f_{start}, f_{end} = FNN_{start}(f_a), FNN_{end}(f_a)\\
&s_{mask} = FNN_{mask}(f_a)
\end{eqnarray}
where $w_a \in \mathbb{R}^{T}$ is the attentive weights, which adaptively measures the importance score of each frame. $FNN_{AP}, FNN_{start}, FNN_{end}$, and $FNN_{mask}$ are separate feedforward neural networks that retain the original dimensions of the input.

\subsection{Clip-level Predictor (CLP)}
The proposed CLP keeps the same design philosophy as previous clip-level methods, which derive the clip-level feature map by aggregating specific frames according to the coordinate. We don't realize this operation by stacking the convolution layers, instead by introducing the sampling mask weight $W$ proposed in \cite{lin2019bmn}. Moreover, the generation-guided architecture is also applied to alleviate the gap between visual and semantic features.

In general, a Dense Proposal Generation Module (DPGM) and Clip-Level Generator (CLG) are first employed to generate the visual feature map $M_v$ and implicit clip feature $f_{c}$, respectively. Then $f_{c}$ is compared with the short-cut $M_v$ by Hadamard product following $LateConv$ to derive the clip-level score map ($S_c$). $LateConv$ consists of several layers of convolution. This process can be roughly expressed as:

\begin{eqnarray}
&M_v = DPGM(V_f)\\
&f_{c} = CLG(M_f) \\
&S_c = LateConv(M_v \odot f_{c})
\end{eqnarray}

\subsubsection{\textbf{Dense Proposal Generation Module (DPGM)}}

\begin{figure}[ht]
\begin{center}
\includegraphics[scale=0.20]{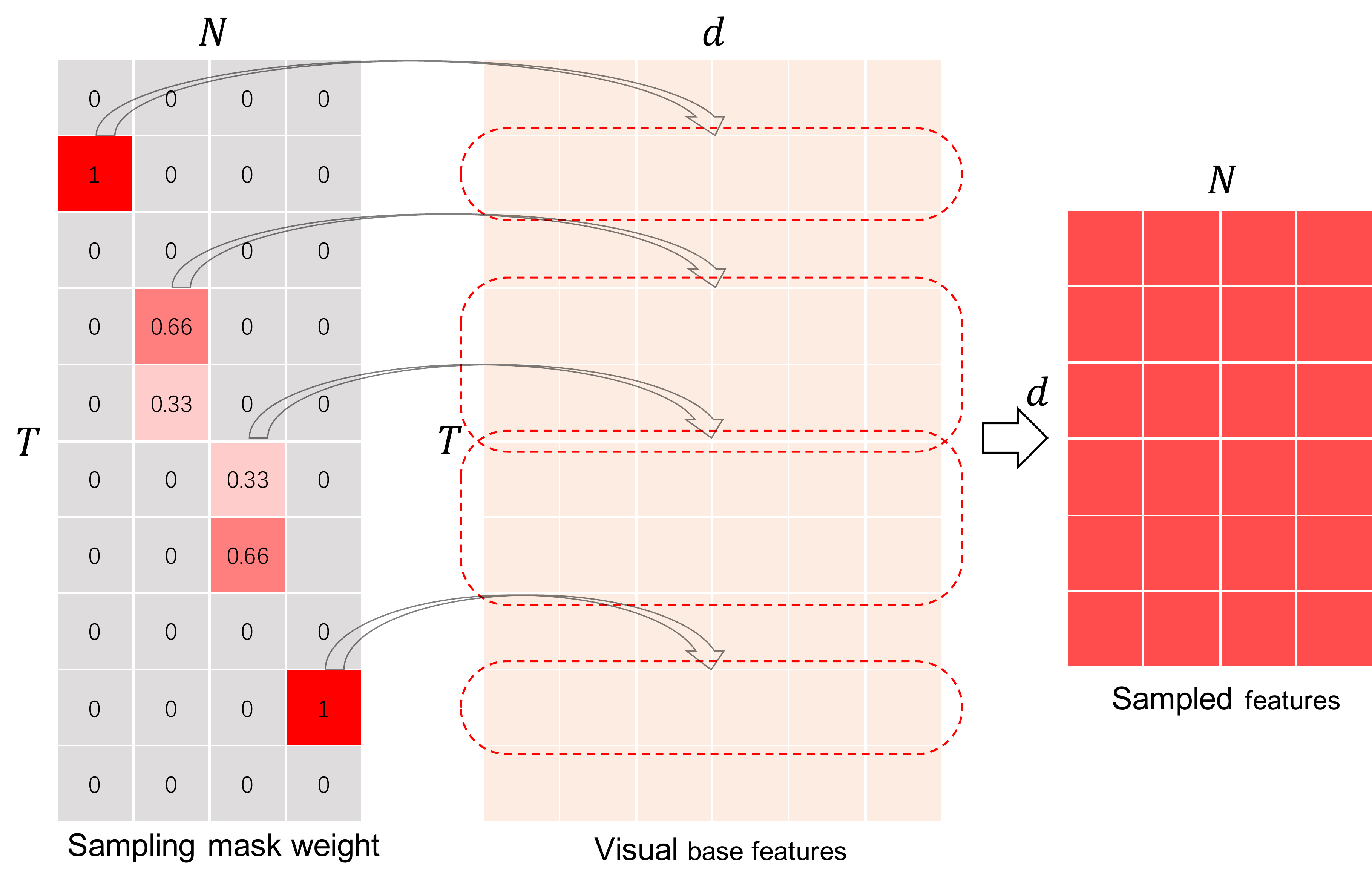}
\end{center}
\caption{The diagram of the dense sampling in DPGM. We adopt the sampling mask weight $W$ proposed in \cite{lin2019bmn} to perform the matrix multiplication with visual base features. The nonzero value $w_{i,j}$ in $W$ is annotated in red, which denotes to sample the $i$-th feature with a weight of $w_{i,j}$.}
\label{DSO}
\end{figure}

As shown in Fig. \ref{DSO}, the matrix multiplication of sampling mask weight $W \in \mathbb{R}^{N \times T \times T \times T}$ and visual base feature $V_f \in \mathbb{R}^{d \times T} $ generates the sampled proposal feature map $S_m \in \mathbb{R}^{N \times d \times T \times T} $, where $N$ denotes the sampling number. It's worth noting that there are three dimensions equal to $T$ in $W$, it is because, for each start time and end time, there is a weight $w \in \mathbb{R}^{T \times N}$. We subsequently adopt the Maximum pooling operation towards $S_m$ at $N$ dim and obtain the visual feature map $M_v \in \mathbb{R}^{d \times T\times T}$. 

\begin{eqnarray}
&M_v = Max( W \times V_f, dim\ =\ 0)
\end{eqnarray}
Where $Max(\ \cdot, dim\ =\ 0)$ denotes calculating the maximum value at $N$ dim.

\subsubsection{\textbf{Clip-level Generator (CLG)}}
\label{3.4.2}
This module aims to derive the implicit clip feature $f_{c}$, which owns a narrower gap with visual information than the semantic one. We first utilize $EarlyConv$ to increase the receptive fields and then fuse with semantic feature $s$ by Hadamard product to exploit the text information.
\begin{eqnarray}
&M_f = EarlyConv(M_v) \odot s
\end{eqnarray}
Where $M_f \in \mathbb{R}^{T \times T \times d}$ denotes the fused feature map. Considering the GT in $M_f$ should reach the maximum for all values at dimension $d$. We adopt the $MaxPool2d$ to eliminate the redundant features.
\begin{eqnarray}
&f_{c} = MaxPool2d(M_f)
\end{eqnarray}

\subsection{Loss Function}
For each video and its query, GMU outputs start scores $s_{start}$, end scores $s_{end}$, masked token features $s_{mask}$, attentive weights $w_a$, and clip-level score map $M_c$. $s_{start}$ and $s_{end}$ correspond to boundary location loss $\mathcal{L}_{BL}$ \cite{lin2019bmn}. The others correspond to Cross-Entropy (CE) loss $\mathcal{L}_{CE}$ \cite{1999The}, temporal attention guidance (TAG) loss $\mathcal{L}_{TAG}$ \cite{mun2020local}, and binary cross entropy (BCE) loss $\mathcal{L}_{BCE}$ \cite{zhang2020learning}, respectively. The total loss consists of four parts.

\begin{eqnarray}
\mathcal{L} = \mathcal{L}_{BL} + \mathcal{L}_{CE} + \mathcal{L}_{TAG} + \mathcal{L}_{BCE}
\end{eqnarray}

\subsubsection{\textbf{Boundary Location Loss.}}
$s_{start}^i, s_{end}^i$ denote the probability of the $i$-th moment $m_i$ covering the start or end timestamps. Note that $m_i$ covers the period $p_i = D_m[i-0.5, i+0.5]$, where $D_m = \frac{D}{T}$ is the duration of each moment. Considering the ambiguous boundary definition, we expand the duration of the start and end moments. We utilize the IoU of $p_i$ and $[t_s-1.5D_m, t_s+1.5D_m]$ or $[t_e-1.5D_m, t_e+1.5D_m]$ to be the label, where $t_s$ and $t_e$ denote the start and end timestamps.
\begin{eqnarray}
\mathcal{L}_{BL} = \mathcal{L}_{wbl}(s_{start}, l_{start})+\mathcal{L}_{wbl}(s_{end}, l_{end})
\end{eqnarray}
where $l_{start}$ and $l_{end}$ denote the start and end labels .$\mathcal{L}_{wbl}$ denotes the weighted binary logistic regression loss function \cite{lin2019bmn}.
\begin{eqnarray}
\mathcal{L}_{wbl}(s,l) = -\frac{1}{N_b}\sum_{i=1}^{N_b} \alpha^+ b_i \rm{log}s_i + \alpha^- (1-b_i) \rm{log}(1-s_i)
\end{eqnarray}
Where $b_i = sign(l_i-\theta)$ and $N_b = B \cdot T$ denotes the moment amount in a batch $B$. $\theta = 0.5$ is a threshold. $\alpha^+ = \frac{N_b}{\sum b_i}$, $\alpha^- = \frac{N_b}{N_b-\sum b_i}$. They denote the positive and negative weights, respectively.

\subsubsection{\textbf{Cross-Entropy loss.}}
We regard the Masked Language Modeling (MLM) task as a classification problem and adopt the CE loss to solve it. It can be expressed as:
$\mathcal{L}_{CE} = \frac{1}{B}\sum_{i=1}^B \rm{log}s_{i,I_i}$,
where $I_i$ denotes the index of the masked token.

\subsubsection{\textbf{Temporal Attention Guidance loss.}}
$w_a$ reflects the importance score of each frame, which should be larger for the moments within the ground truth. We adopt the TAG loss \cite{yuan2019find} to lead the attentive pooling operation to learn balanced scores. It can be expressed as:
$\mathcal{L}_{TAG} = -\sum_{i=1}^B \frac{\sum_{j=1}^T \hat{w}_a^{i,j} \rm{log}{w_a^{i,j}}}{\sum_{j=1}^T \hat{w}_a^{i,j}}$
, where $\hat{w}_a^{i,j}$ is a binary score measuring if the $j$-th moment in $i$-th sample is within the ground truth.

\subsubsection{\textbf{Binary Cross Entropy loss}} 
Same as previous clip-based methods \cite{zhang2020learning}, we adopt the BCE loss with scaled IoU scores $y$ as the label to conduct the training of CLP.
\begin{eqnarray}
y_i=\left\{
\begin{aligned} 0 \qquad \qquad o_i \leq o_{min} \\
\frac{o_i-o_{min}}{o_{max}-o_{min}} \quad Otherwise\\
\quad 1 \qquad \qquad o_i \geq o_{max} 
\label{eqeq}
\end{aligned}
\right.
\end{eqnarray}
Where $o_i$ denotes the original IoU score of $i$-th clip candidate. $o_{max}$, $o_{min}$ are two thresholds. Consequently, BCE loss can be expressed as:
$\mathcal{L}_{BCE} = \frac{1}{N_c}\sum\limits_{i=1}^{N_c} y_i \rm{log} p_i + (1-y_i)\rm{log} (1-p_i)$
where $N_c = B\cdot T \cdot T$ is the total number of candidates in a batch. $p$ denotes the predicted score in the clip-level score map $M_c$.

\section{Experiments}
\subsection{Datasets}
\textbf{Charades-STA \cite{sigurdsson2016hollywood}.} It contains 6,672 indoor videos ranging from 2.4 to 194.4 seconds with 16,128 moments, whose timestamps are annotated precisely. The average duration of the untrimmed videos is less than 30 seconds, and the average duration of labeled moments, it's 8.2 seconds. So it's a small dataset with short videos. For fair comparisons, we follow the same split as \cite{mun2020local,zhang2020learning}. There are 12,408 moments for training and 3,720 moments for the test. 

\textbf{ActivityNet Caption \cite{krishna2017dense}.} 
It contains 20K untrimmed videos ranging from 1.1 to 975.0 seconds, which is a vast duration span. Each video is annotated with 3.65 queries on average, and each query is comprised of 13.48 words on average. The official split divides it with a 2:1:1 ratio. There are 37,421, 17,505, 15,031 moments in the training set, val\_1 split, and val\_2 split. Following \cite{zhang2020learning}, we regard the val\_2 split as the test set. Overall, it's a larger dataset with longer videos.
    

\subsection{Implementation details}
\subsubsection{\textbf{Metrics.}} 
We adopt the indicator "R@$n$, IoU=$m$" and mIoU to report the evaluation. We set $n=1$ and $m\in\{0.1, 0.3, 0.5, 0.7\}$. Some methods \cite{zhang2020learning} also report the results for $n=5$, which is significantly affected by the threshold of Non maximum suppression (NMS) \cite{neubeck2006efficient} during inference. A larger "R@5, IoU=0.5" may mean a smaller "R@5, IoU=0.7". It is unfair to conduct the comparisons in the case that the hyper-parameter in the post-process is different. So we concentrate on the R@1 and mIoU.

\subsubsection{\textbf{Implementation Details.}}
On Charades-STA, we conduct experiments with VGG, C3D, and I3D as the pre-trained 3D-CNN to perform a fair comparison. As for AvtivityNet captions, we adopt the C3D and I3D features. Clip number $T$ and dimension $d$ is set to be 64 and 512 for both Charades-STA and AvtivityNet captions. All experiments are conducted on the NVIDIA V100 WORKSTATION with Pytorch \cite{paszke2019pytorch} implementation. The optimizer is Adam \cite{kingma2014adam} with a decay factor of 0.5. Learning rate equals to $1\times10^{-4}$, batch size $B$ is fixed to be 32. The thresholds $o_{max}$ and $o_{min}$ are set to be 0.5 and 1.0 for Charades-STA and ActivityNet Captions. During inference, we report the results of the moment-level score map $S_m$, clip-level score map $S_c$, and fusion score map $S_f = S_m \odot S_c$. $n_l$ and $n_g$ are set to 3 and 2 as \cite{mun2020local}. The $EarlyConv$ in Sec. \ref{3.4.2} consists of 8 layers of convolution with input channels, output channels, kernel size, stride, padding, and dilation equal to \{512, 512, 5, 1, 16, 1\} for the first layer and \{512, 512, 5, 1, 0, 1\} for the last 7 layers. As for the $LateConv$ comprised of 9 layers of convolution, its first 8 layers are set to be the same as $EarlyConv$. The final layer is set to \{512, 1, 1, 1, 0, 0\}. Because the clip-level converges faster than the moment-level as shown in Fig. \ref{AB_fusion}, we set the weight of $\mathcal{L}_{BCE}$ to be 0.001 after $upsilon$ epoch. $upsilon$ equals 9 and 7 on Charades-STA and ActivityNet caption.


\subsection{Comparisons with previous SOTA}
\label{result_compare}

\begin{table}[h]
\centering
\caption{The comparisons with previous methods on Charades-STA. Note that LGI and 2D-TAN are GMU's moment-level and clip-level baseline, respectively. GMU-M, -C, -F denote the results derived from moment-level, clip-level, and fusion score maps, respectively. "M", "C", and "F" represents the moment, clip, and fusion level separately.}

\label{charades_result}
\setlength{\tabcolsep}{0.9mm}{
\begin{tabular}{c|c|c|c|c|c|c}
\hline
\multirow{2}*{3D CNN}&\multirow{2}*{Level}&\multirow{2}*{Method}&\multicolumn{3}{c|}{R@1, IoU=} & \multirow{2}*{mIoU}\\
& & &\multicolumn{3}{c|}{\ 0.3 \quad\ \ \ 0.5\quad\ \ \ \  0.7\ } &\\
\hline
\hline

\multirow{12}*{VGG}
&\multirow{5}*{M}

&LGI \cite{mun2020local} &- &41.72 &21.48 &- \\
&&CBLN \cite{liu2021context}&- &43.67 &24.44 &-\\
&&CPN \cite{zhao2021cascaded} &\textbf{64.41} &{46.08} & 25.06 &\uline{43.90}\\
&&SSCS \cite{ding2021support}&- &43.15 &25.54 &-\\
&&GMU-M (ours) &\uline{61.72} &44.49 &24.57 &42.12 \\
\cline{2-7}

&\multirow{4}*{C}
&2D-TAN \cite{zhang2020learning} &-&39.70 &23.31 &- \\
&&FVMR \cite{gao2021fast} &- &42.36 &24.14 &- \\
&&MS-2D-TAN \cite{zhang2021multi} &- &45.65 &\uline{27.20} &-\\
&&GMU-C (ours) &60.75 & {46.18}&26.16 &41.83\\
\cline{2-7}

&\multirow{3}*{F}
&APGN \cite{liu2021adaptive} &- &44.23 &25.64 &-\\
&&DPIN \cite{10.1145/3394171.3413975} &- &\uline{47.98} &26.96 &-\\
&&GMU-F (ours) &\textbf{64.41} &\textbf{48.76} &\textbf{27.72} &\textbf{44.22} \\
\hline
\hline
\cline{2-7}

\multirow{10}*{C3D} 
&\multirow{3}*{M}
&CTRL \cite{gao2017tall}&-&23.63&8.89&-\\
&&CBP \cite{wang2020temporally}&- &36.80 &18.87  &35.74 \\
&&GMU-M (ours) &{56.61} &39.33 &19.01 &37.69 \\

\cline{2-7}
&\multirow{3}*{C}
&FVMR \cite{gao2021fast}&- &38.16 &18.22 &- \\
&&MS-2D-TAN \cite{zhang2021multi}&- &41.10 &\textbf{23.25} &- \\
&&GMU-C (ours) &55.67 &\uline{41.85} &22.63 &{38.16} \\

\cline{2-7}
&\multirow{3}*{F}
&BPNet \cite{xiao2021boundary} &55.46 &38.25 &20.51 &38.03 \\
&&LPNet \cite{xiao2021natural} &\textbf{59.14} &40.94 &21.13 &\uline{39.67} \\
&&GMU-F (ours) &\uline{58.28} &\textbf{43.39} &\uline{23.17} &\textbf{39.74} \\
\hline
\hline

\multirow{9}*{I3D}
&\multirow{3}*{M}

&LGI \cite{mun2020local}&\textbf{72.96} & 59.46 & 35.48 &\uline{51.28}\\
&&SSCS \cite{ding2021support}&- &\uline{60.75} &36.19 &- \\
&&GMU-M (ours) &72.02 &56.51 &32.85 &50.19 \\
\cline{2-7}
&\multirow{3}*{C}
&2D-TAN \cite{zhang2020learning}&- &50.62 &28.71 &- \\
&&MS-2D-TAN \cite{zhang2021multi}&- &60.08 &\uline{37.39} &- \\
&&GMU-C (ours) &69.84 &57.77 &35.97 &50.06 \\
\cline{2-7}
&\multirow{3}*{F}
&BPNet \cite{xiao2021boundary} &65.48 &50.75 &31.64 &46.34  \\
&&LPNet \cite{xiao2021natural} &66.58 &54.33 &34.03 &47.71 \\
&&GMU-F (ours) &\uline{72.58} &\textbf{61.18} &\textbf{37.45} &\textbf{52.18} \\

\hline
\end{tabular}}
\end{table}

\begin{table}[h]
\centering
\caption{The comparisons with previous methods on ActivityNet Caption. }

\label{anet_result}
\setlength{\tabcolsep}{1mm}{
\begin{tabular}{c|c|c|c|c|c|c}
\hline
\multirow{2}*{3D CNN}&\multirow{2}*{Level}&\multirow{2}*{Method}&\multicolumn{3}{c|}{R@1, IoU=} & \multirow{2}*{mIoU}\\
&& &\multicolumn{3}{c|}{\ 0.3 \quad\ \ \ 0.5\quad\ \ \ \  0.7\ } &\\
\hline
\hline
\multirow{14}*{C3D}

&\multirow{5}*{M}
&DRN \cite{zeng2020dense}&- &45.45 &24.39 &-\\
&&LGI \cite{mun2020local}&58.52 &41.51 &23.07 &41.13\\

&&SSCS \cite{ding2021support}&61.35 &{46.67} &27.56 &-\\
&&IVG-DCL \cite{nan2021interventional} &{63.22} &43.84 &27.10 &44.21\\
&&GMU-M (ours) &63.00 &43.56 &24.67 &44.32 \\

\cline{2-7}
&\multirow{5}*{C}
&2D-TAN \cite{zhang2020learning} &59.45 &44.51 &26.54 &-\\
&&CMIN  \cite{zhang2019cross} &\uline{63.61} &43.40 &23.88 &-\\
&&FVMR \cite{gao2021fast}&60.63 &45.00 &26.85 &- \\
&&MS-2D-TAN \cite{zhang2021multi}&61.04 &46.16  &\textbf{29.21} &- \\
&&GMU-C (ours) &61.58 &\uline{47.04} &\uline{28.98} &\uline{45.11} \\

\cline{2-7}
&\multirow{4}*{F}
&ReLoCLNet \cite{zhang2021video} &42.65 &28.54 &17.76&-\\
&&BPNet \cite{xiao2021boundary} &58.98 &42.07 &24.69 &42.11 \\
&&LPNet \cite{xiao2021natural} &\textbf{64.29} &45.92 &25.39 &44.72 \\
&&GMU-F (ours) &{62.36} &\textbf{47.08} &{28.61} &\textbf{45.48} \\
\hline

\multirow{7}*{I3D}

&\multirow{3}*{M}

&ExCL \cite{ghosh2019excl} &62.30 &42.70 &24.10&-\\
&&VSLNet \cite{zhang2020span} &63.16 &43.22 &26.16 &-\\
&&GMU-M (ours) &\textbf{63.17} &43.53 &26.80 &44.17 \\

\cline{2-7}
&\multirow{3}*{C}
&TMLGA \cite{rodriguez2020proposal}&51.28 &33.04 &19.26 &-\\
&&MS-2D-TAN \cite{zhang2021multi} &62.09 &45.50 &28.28 &-\\
&&GMU-C (ours) &62.20 &\uline{46.11} &\uline{28.37} &\uline{45.01} \\

\cline{2-7}
&\multirow{1}*{F}
&GMU-F (ours) &\uline{63.06} &\textbf{46.48} &\textbf{28.70} &\textbf{45.39} \\
\hline
\end{tabular}}
\end{table}






We compare the performance of GMU's every level with previous SOTA as shown in Table \ref{charades_result} and Table \ref{anet_result}.
For each kind of 3D CNN features on the two benchmarks, the proposed GMU reaches SOTA.

On Charades-STA, for the fusion map of GMU (GMU-F) with VGG features, it surpasses all the previous SOTA with an increment of 0.78\% at IoU=0.5 and 0.52\% at IoU=0.7. Moreover, the separate moment-level and clip-level results significantly outperform the baselines. GMU-M realizes a boost of 2.77\% at IoU=0.5 than LGI, and it turns to 4.46\% for GMU-C compared with 2D-TAN. If adopting the C3D features, the gain at IoU=0.7 and mIoU expands to 3.33\% and 2.52\%. As for the I3D features, GMU-F reaches SOTA at IoU=0.5, IoU=0.7, and mIoU. GMU-C significantly outperforms 2D-TAN with 7.15\% at IoU=0.5 and 4.14\% at IoU=0.7. In this benchmark, fusing the two levels totally improves the performance at all metrics.

As for ActivityNet caption with C3D features, the proposed GMU reaches SOTA at IoU=0.5 and mIoU. The individual clip-level map displays great art and realizes the second-highest performance at three indicators. It surpasses 2D-TAN with 2.53\% at IoU=0.5 and 2.44\% at mIoU. GMU-M also outperforms LGI with 4.48\% at IoU=0.3 and 2.05\% at mIoU=0.5. With I3D features, GMU-F reaches SOTA at IoU=0.5, 0.7, and mIoU, while GMU-C ranks second. The moment-level still performs best at IoU=0.3, and it is reduced 0.11\% when fusing with the clip-level score map.

The proposed unification strategy of training significantly boosts performance at all levels. Pure moment-level and clip-level both surpass their corresponding baselines. From the distributional analysis of the results, the moment-level always exceeds the clip-level at IoU=0.3, and it is on the opposite at IoU=0.5 and IoU=0.7. The advantages of the moment-level lie in overall judgments from the up-down perspective. Accordingly, its recall rate performs better at coarse grounding. The clip-level predicts from the bottom-up perspective and outperforms the other at fine-grained grounding. A unification of them combines their merits and reaches a balance at all level of precision, leading to a permanent improvement at mIoU. 

Moreover, we also compare GMU with other approaches based on multi-level fusion, and the results remain competitive. GMU-F significantly outperforms previous fusion methods in the exhibited five experiments. The advantages of GMU mainly reflect in two points: 1. It adopts more independent, effective, and specialized multi-level separate predictors with MLM task and Dense sampling. 2. We leverage the strategy of "training separately, predicting together.".

We also conduct experiments on TACOS \cite{regneri-etal-2013-grounding} to prove the generalization of GMU. Please refer to Table 1 in the appendix.

\section{Ablation Study}
\begin{figure}[h]
\begin{center}
\subfigure[The mIoU of each map in terms of the training epoch.]{\includegraphics[scale=0.25]{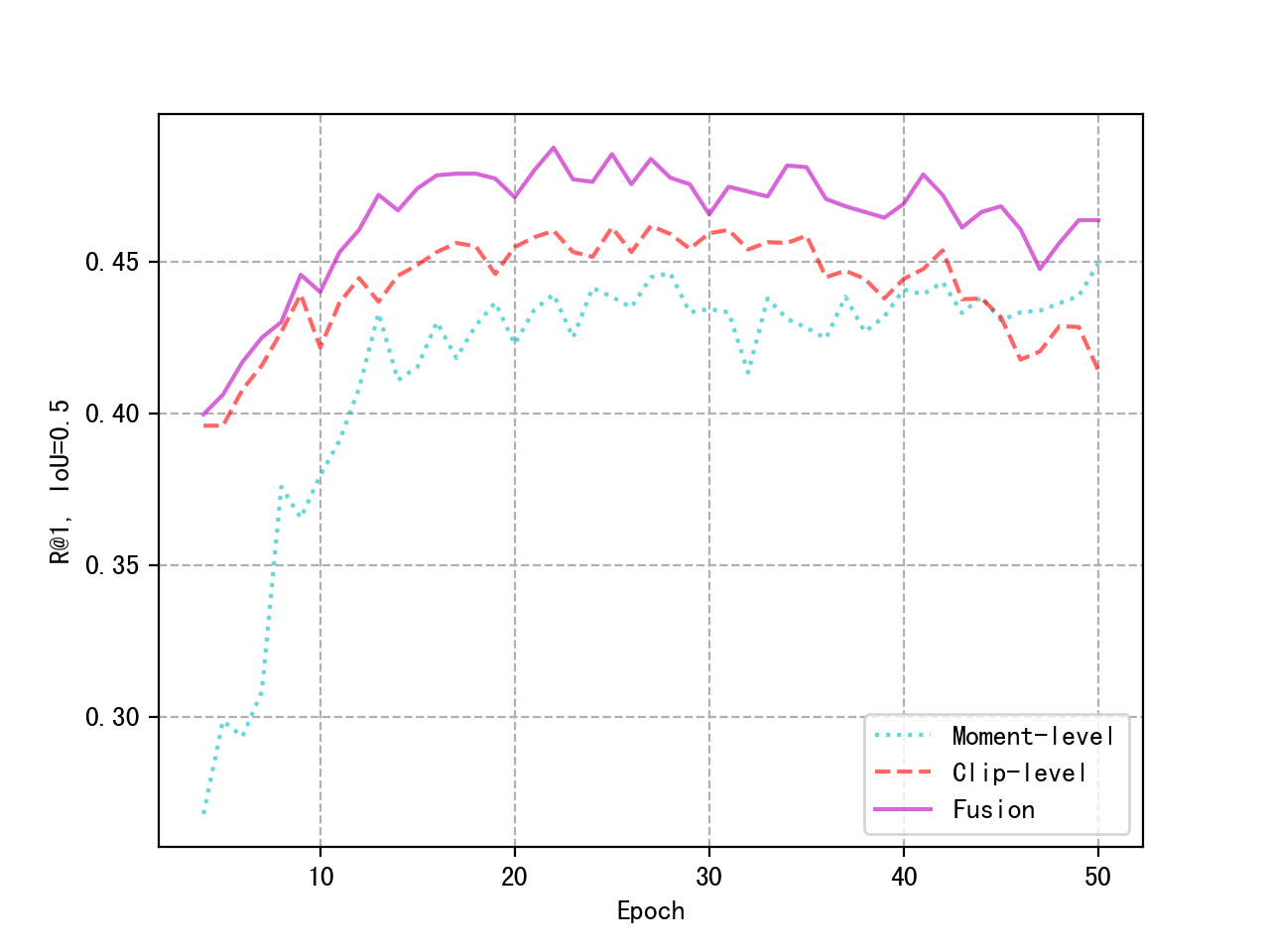}\label{AB_a}}
\subfigure[The "R@1, IoU=0.5" of each map in terms of the training epoch.]{\includegraphics[scale=0.25]{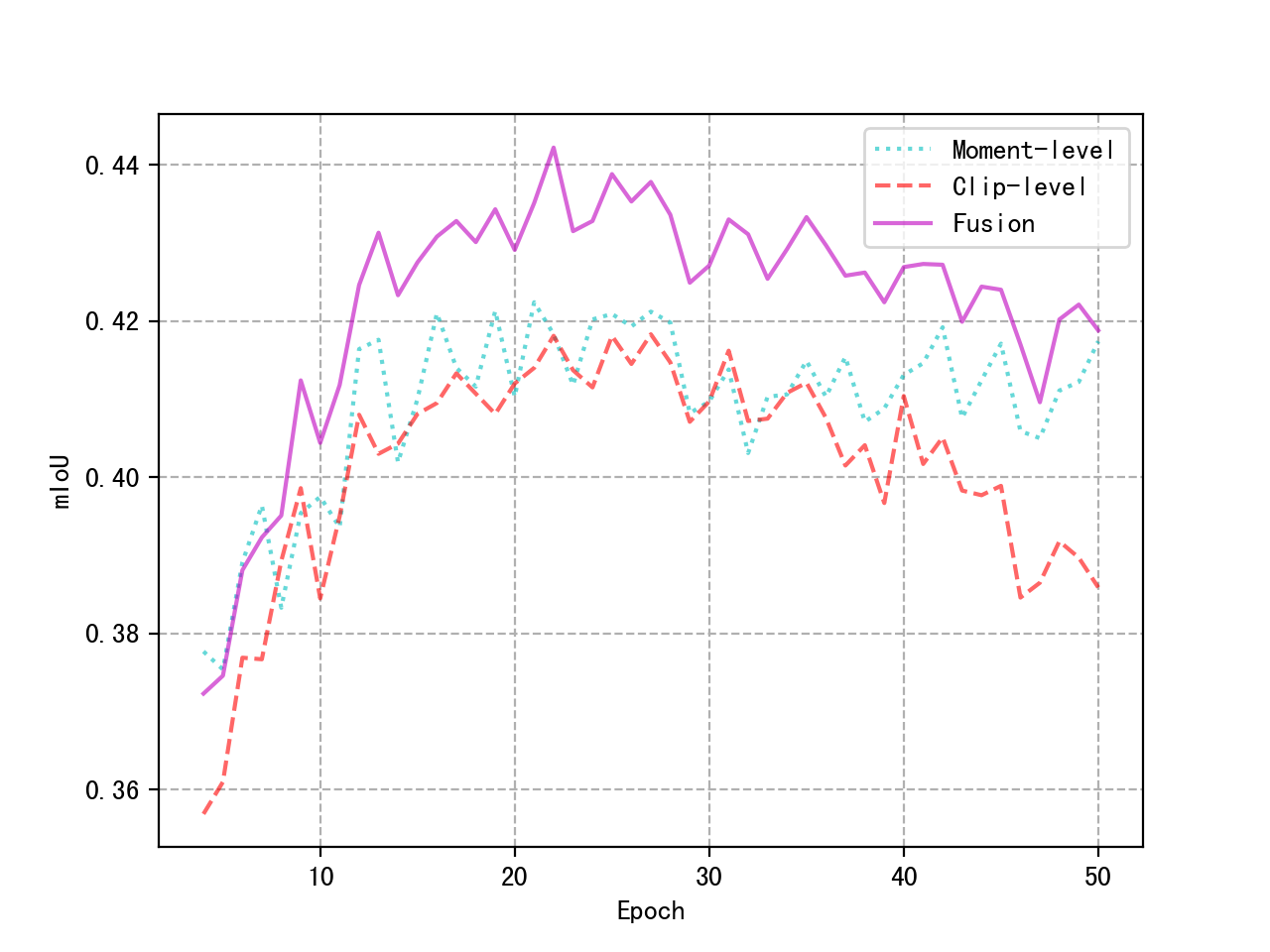}\label{AB_b}}
\end{center}
\caption{The performance comparison of the generated maps in terms of the training epoch on Charades-STA test set.}
\label{AB_fusion}
\end{figure}

The result of the ablation study is shown in Fig.\ref{AB_fusion}, Table \ref{AB_other}, Table \ref{AB_loss}, and Table \ref{AB_CLP}.

\textbf{Multi-level unified strategy.} Fig. \ref{AB_fusion} investigates the variations of each map's performance towards the training epoch. Fusion map methods always surpass the separate moment-level or clip-level ones, which indicates the success of the multi-level unified strategy. More importantly, the comparisons of row 7 and row8, row 9 and row 19, row 10 and row 20, row 22 and row 25, row 23 and row 26 in Table \ref{AB_other} demonstrate that Sharing Encoder (SE) increases the performance of each level. A huge increment is realized for clip-level with around 3\%-5\%, indicating the superiority of unifying both levels also embodies in learning better base encoders, including the visual base module and semantic encoder.
Moreover, the performance of clip-level in Fig. \ref{AB_a} and Fig. \ref{AB_b} declines after some epochs of training, while moment-level retains to fluctuate around a stable value. Accordingly, it's necessary to decrease the weight of $\mathcal{L}_{BCE}$ after $\upsilon$ epochs.
Row 27-30 in Table \ref{AB_other} explores its impacts. 

\begin{table}[h]
\centering
\caption{The ablation study of the relevant strategy and hyper-parameters. SE denotes the strategy of sharing encoders, including the visual base module and semantic encoder. GG denotes the generation-guided strategy. $\kappa_e$ and $\kappa_l$ denote the convolution layer numbers of the $EarlyConv$ and $LateConv$. $\upsilon$ denotes the transition epoch after which we set a weight of 0.001 for the loss of the CLP part. $N$ denotes the sampling number in $W$.}
\label{AB_other}
\setlength{\tabcolsep}{1.0mm}{
\begin{tabular}{c|c|cc|ccccc|cc}
\hline
\multirow{2}*{Level}&\multirow{2}*{Row}&\multicolumn{2}{c|}{Strategy} &\multicolumn{5}{c|}{Hyper-parameters}& R@1 & \multirow{2}*{mIoU}\\

& & SE &GG &$\kappa_e$& $\kappa_l$ &$\upsilon$& $T$ & $N$  & IoU=0.5 &\\
\hline
\hline
\multirow{8}*{\makecell[c]{Moment}}
&1&\XSolidBrush &\XSolidBrush &-&- &- &16 &- &38.01 &36.76\\
&2&\XSolidBrush &\XSolidBrush &-&- &- &32 &- &39.27 &37.11\\
&3&\XSolidBrush &\XSolidBrush &-&- &- &64 &- &39.81 &37.56\\
&4&\XSolidBrush &\Checkmark &-&- &- &8 &- &36.83 &36.52\\
&5&\XSolidBrush &\Checkmark &-&- &- &16 &- &38.92 &37.02\\
&6&\XSolidBrush &\Checkmark &-&- &- &32 &- &40.01 &37.62\\
&7&\XSolidBrush &\Checkmark &-&- &- &64 &- &40.23 &37.87\\
&8&\Checkmark &\Checkmark &-&- &- &64 &- &\textbf{44.49} &\textbf{42.12}\\
\hline
\multirow{13}*{Clip} 
&9&\XSolidBrush &\XSolidBrush &-&9 &- &64 &4 &42.71 &39.21\\
&10&\XSolidBrush &\XSolidBrush &-&17 &- &64 &4 &42.60 &38.98\\

&11&\XSolidBrush &\Checkmark &8&9 &- &8 &4 &40.01 &37.28\\
&12&\XSolidBrush &\Checkmark &8&9 &- &16 &4 &41.32 &38.40\\
&13&\XSolidBrush &\Checkmark &8&9 &- &32 &4 &41.76 &38.88\\
&14&\XSolidBrush &\Checkmark &8&9 &- &64 &4 &42.53 &39.07\\

&15&\XSolidBrush &\Checkmark &8&9 &- &64 &2 &41.39 &38.41\\
&16&\XSolidBrush &\Checkmark &8&9 &- &64 &8 &42.49 &39.12\\
&17&\XSolidBrush &\Checkmark &8&9 &- &64 &16 &42.58 &39.16\\
&18&\XSolidBrush &\Checkmark &8&9 &- &64 &32 &42.74 &39.47\\

&19&\Checkmark &\XSolidBrush &-&9 &- &64 &4 &45.88 &41.63\\
&20&\Checkmark &\XSolidBrush &-&17 &- &64 &4 &45.52 &41.59\\
&21&\Checkmark &\Checkmark &8&9 &- &64 &4 &\textbf{46.18} &\textbf{41.83}\\
\hline
\multirow{9}*{Fusion} 
&22&\XSolidBrush &\XSolidBrush &-&9 &10 &64 &4 &43.82 &40.79 \\
&23&\XSolidBrush &\XSolidBrush &-&17 &10 &64 &4 &44.09 &40.92 \\
&24&\XSolidBrush &\Checkmark &8&9 &10 &64 &4 &44.81 &41.58\\
&25&\Checkmark &\XSolidBrush &-&9 &10 &64 &4 &47.69 &43.74\\
&26&\Checkmark &\XSolidBrush&-&17 &10 &64 &4 &47.54 &43.26 \\
&27&\Checkmark &\Checkmark &8&9 &10 &64 &4 &\textbf{48.76} &\textbf{44.22}\\
&28&\Checkmark &\Checkmark &8&9 &9 &64 &4 &48.23 &43.92\\
&29&\Checkmark &\Checkmark &8&9 &11 &64 &4 &48.20 &43.73\\
&30&\Checkmark &\Checkmark &8&9 &\XSolidBrush &64 &4 &47.34 &43.03\\
\hline
\end{tabular}}
\end{table}

\textbf{Generation-guided strategy and the hyper-parameters.} 
As shown in Table \ref{AB_other}, the comparison of row 1-3 with row 5-7 proves that generation-guided strategy improves the performance at moment-level with various configurations. The fusion one also relies on it much to further improve according to row 23 and row 24, row 26, and row 27. The novel framework innovates in replacing the original semantic feature with the generated feature, which reduces the gap between query and gallery. Moreover, the shortcut connection from the original visual features reinforces the memory of visual information. We also test the effects of the number of pre-segmented moments $T$, a larger $T$ is significantly beneficial for both levels according to row 1-7 and row 11-14. As for the sampling number $N$, the profit caused by a larger $N$ is slight referring to row 14-18 if $N > 4$. For the balance of the performance and time cost, we keep $N$ equal to 4 during training.

\textbf{Loss function.} Each loss term plays a crucial part referring to Table \ref{AB_loss}. For the moment-level, the model with pure boundary location loss $\mathcal{L}_{BL}$ performs poorly. The elaborate auxiliary MLM task with $\mathcal{L}_{BL}$ keeps the model retaining semantic information and leads to an increment of 3.11\% at "R@1, IoU=0.5". The temporal attention guidance loss $\mathcal{L}_{TAG}$ guides the $NLBlock$ in the moment-level part attention to the moments within ground truth. It boosts the "R@1, IoU=0.5" and mIoU by 8.04\% and 3.64\%. The combination of $\mathcal{L}_{BL}$, $\mathcal{L}_{BL}$, and $\mathcal{L}_{TAG}$ achieves the best moment-level performance. Besides, the fusion one with the four loss terms surpasses all other combinations. Compared to the separate two levels, it shows comprehensive supremacy.

\begin{table}[h]
\centering
\caption{The ablation study of the loss function.}
\label{AB_loss}
\setlength{\tabcolsep}{1.8mm}{
\begin{tabular}{c|cccc|cc}
\hline
\multirow{2}*{Method}&\multicolumn{4}{c|}{Loss terms}& R@1 & \multirow{2}*{mIoU}\\
&$\mathcal{L}_{BL}$ &
$\mathcal{L}_{CE}$ &
$\mathcal{L}_{TAG}$ &
$\mathcal{L}_{BCE}$ &IoU=0.5 &\\
\hline
\hline
\multirow{4}*{\makecell[c]{Moment-\\level}}& \Checkmark&&& &31.80 &34.06\\
&\Checkmark&\Checkmark&&&34.92&35.33\\
&\Checkmark&&\Checkmark&&39.84&37.70\\
&\Checkmark&\Checkmark&\Checkmark&&40.23 &37.87\\
\hline
\multirow{1}*{Clip-level}&&&&\Checkmark &42.53 &39.07\\
\hline
\multirow{3}*{Fusion}
&\Checkmark&&&\Checkmark &44.67 &41.78\\
&\Checkmark&&\Checkmark&\Checkmark &47.01 &42.89 \\
&\Checkmark&\Checkmark&\Checkmark&\Checkmark &48.76 &44.22\\
\hline
\end{tabular}}
\end{table}

\textbf{The components of the CLP.} The results in Sec. \ref{result_compare} exhibit a significant improvement of CLP than clip-level baseline 2D-TAN. The major modifications are DPGM, GG, and SE referring to Table \ref{AB_CLP}. 2D-TAN aggregates all the frames within a proposal and predicts sparsely. We sample a fixed number of frames and predict densely. With only DPGM, CLP outperforms 2D-TAN, which conforms to the analysis for row 14-18 in Table \ref{AB_other}. Sampling more frames or all frames doesn't bring increments, while densely predicting for all proposals does. Moreover, if adopting the sharing encoder (SE) and generation-guided (GG) strategy, the individual 2D-TAN and the proposed CLP are both boosted, which indicates their superiority. 

\begin{table}[h]
\centering
\caption{The ablation study of the clip-level configurations. SC and MP denote the stacked convolution and the max pooling, they are adopted by 2D-TAN as visual aggregation strategies. GG and SE are explained in Table \ref{AB_other}.}
\label{AB_CLP}
\setlength{\tabcolsep}{1.8mm}{
\begin{tabular}{c|ccccc|cc}
\hline
\multirow{2}*{Method}&\multicolumn{5}{c|}{Components}&
\multicolumn{2}{c}{R@1, IoU=}\\
&SC &MP  &DPGM & GG &SE
 &0.5 &0.7\\
\hline
\hline
\multirow{4}*{\makecell[c]{2D-TAN\\(baseline)}}& \Checkmark&&&& &39.70 &23.31\\
&&\Checkmark&&&&39.81 &23.25\\
&\Checkmark&&&\Checkmark&\Checkmark&44.21 &23.44\\
&&\Checkmark&&\Checkmark&\Checkmark&44.73 &23.69\\
\hline
\multirow{4}*{\makecell[c]{Clip-level\\(ours)}}
&&&\Checkmark&& &42.71 &22.36\\
&&&\Checkmark&\Checkmark& &42.53 &22.42\\
&&&\Checkmark&&\Checkmark&45.52 &25.77\\
&&&\Checkmark&\Checkmark&\Checkmark&46.18 &26.16\\
\hline
\end{tabular}}
\end{table}

\section{Further Analysis}
\label{visualization}
\begin{figure}[ht]
\begin{center}
\includegraphics[scale=0.13]{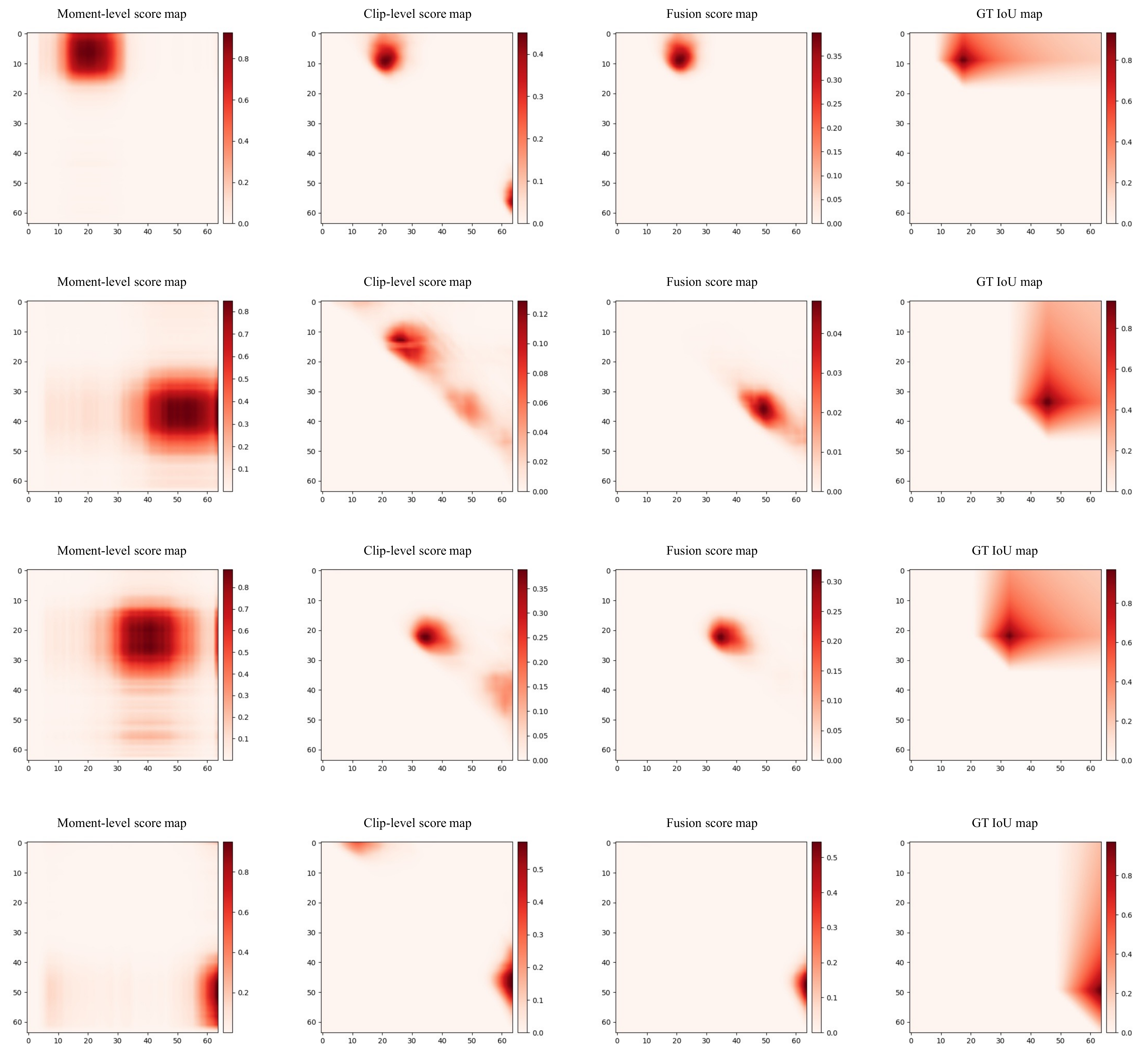}
\end{center}
\caption{The visualization of the score maps and GT IoU map. They are derived from the test set of Charades-STA. Their video name are: "YVKIV", "ZHRPD", "E6DLK", and "FPJ9D".}
\label{heatmap}
\end{figure}

To vividly reflect the impacts of the multi-level unified strategy, we visualize the generated score maps and GT IoU map as shown in Fig. \ref{heatmap}. The first three columns are moment-level, clip-level, and fusion score maps, respectively. The final column is GT IoU maps. We can conclude from the observation: 1) The points of high scores in moment-level score maps intend to be rectangle-like because it is calculated by the biaffine multiplication. This leads it to make blurry judgments towards the unique GT; 2) The distribution of high scores in the clip-level score map is prone to gathering into several clusters around the diagonal. If the top-1 point falls into the wrong cluster, its result is then totally wrong. The relevant phenomenon is explored in \cite{2021}. It's because the CLP directly learns from the overall GT distribution of the benchmark, where GT primarily exists around the diagonal. 3) The GT IoU map always diffuses to the upper right of ground truth. 

Individually, the moment-level one is short in the accurate judgment of the possible boundary; the clip-level one is short in recalling the top-1 point. Consequently, the clip-level "R@1, IoU=0.3" is lower than the moment-level one. As shown in the third column, the fusion score map exactly leverages their merits and avoids their demerits. The redundant points in moment-level one and the wrong clusters in clip-level one are significantly eliminated, so the fusion one achieves the best and balances all metrics. A more detailed exhibition is shown in Fig. 1 in the appendix. 

\section{Conclusion}
In this paper, we propose a novel network named GMU to solve the video grounding task effectively. It unifies the moment-level and clip-level predictors to learn a more robust visual encoder and semantic encoder. The fused map combines its advantages and reaches SOTA performance under low and high precision grounding requirements. Additionally, we pioneer the introduction of a generation-guided paradigm in video grounding. The generated implicit features are utilized to replace the original semantic features and thereby reduce the gap between modals. The short-cut connection also keeps a long-term memory for better retaining visual information. In the future, it's promising to deeply fuse both levels and explore a more flexible generation structure.

\bibliographystyle{ACM-Reference-Format}
\bibliography{sample-base}

\end{document}


\title{Appendix}






\renewcommand{\shortauthors}{Trovato and Tobin, et al.}

\maketitle

\section{Experiments on TACoS}

\subsection{Datasets}
\textbf{TACoS.} It contains 127 videos and 17,344 annotated clips captured in the kitchen. We follow the split as with 50\%, 25\%, and 25\% for training, validation, and test.

\subsection{Comparisons}
As shown in Table \ref{tocos_result}, GMU reaches outstanding performance on TACoS at all metrics. It surpasses MS-2D-TAN by 2.89\% at IoU=0.1 and 1.08\% at IoU=0.3, which further proves the robustness of GMU with multi-level unified strategy. In addition, compared with previous multi-level methods, GMU also presents a more prominent performance. The relevant explanations are given in the main body.

\begin{table}[h]
\centering
\caption{The comparisons with previous methods on TACoS }

\label{tocos_result}
\setlength{\tabcolsep}{2mm}{
\begin{tabular}{c|c|c|c|c|c}
\hline
\multirow{2}*{3D CNN}&\multirow{2}*{Level}&\multirow{2}*{Method}&\multicolumn{3}{c}{R@1, IoU=}\\
&& &\multicolumn{3}{c}{\ 0.1 \quad\ \ \ 0.3\quad\ \ \ \  0.5\ }\\
\hline
\hline
\multirow{13}*{C3D}

&\multirow{6}*{M}
&CBP &- &27.31 &24.79 \\
&&DRN &- &- &23.17 \\
&&CMIN &32.48 &24.64 &18.05 \\

&&IVG-DCL &49.36 &38.84 &29.07 \\
&&CBLN &49.16 &38.98 &27.65 \\
&&GMU-M (ours) &\uline{53.13} &41.27 &28.75\\

\cline{2-6}
&\multirow{4}*{C}
&2D-TAN &47.59 &37.29 &25.32\\
&&FVMR &53.12& 41.48 &29.12 \\
&&MS-2D-TAN &52.39 &\uline{45.61}  &\uline{35.77}\\
&&GMU-C (ours) &52.77 &44.42 &34.69 \\

\cline{2-6}
&\multirow{3}*{F}
&BPNet &- &25.96 &20.96  \\
&&APGN &- &40.47 & 27.86 \\
&&GMU-F (ours) &\textbf{55.28} &\textbf{46.69} &\textbf{35.89} \\
\hline

\end{tabular}}
\end{table}

\section{Prediction Examples}

\begin{figure}[h]
\begin{center}
\subfigure[Video name: ZHRPD. ]{\includegraphics[scale=0.11]{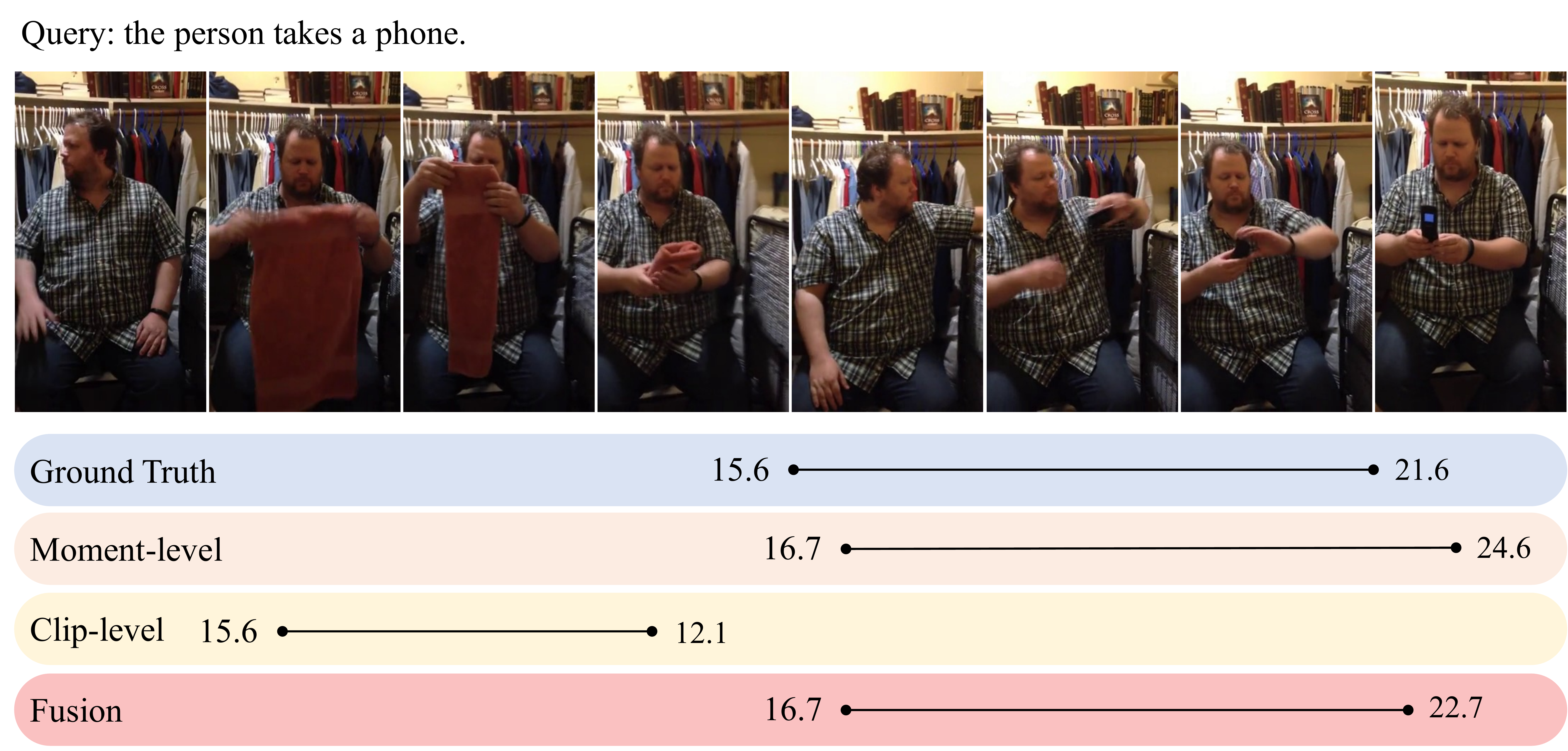}\label{v_i1}}
\subfigure[Video name: E6DLK]{\includegraphics[scale=0.11]{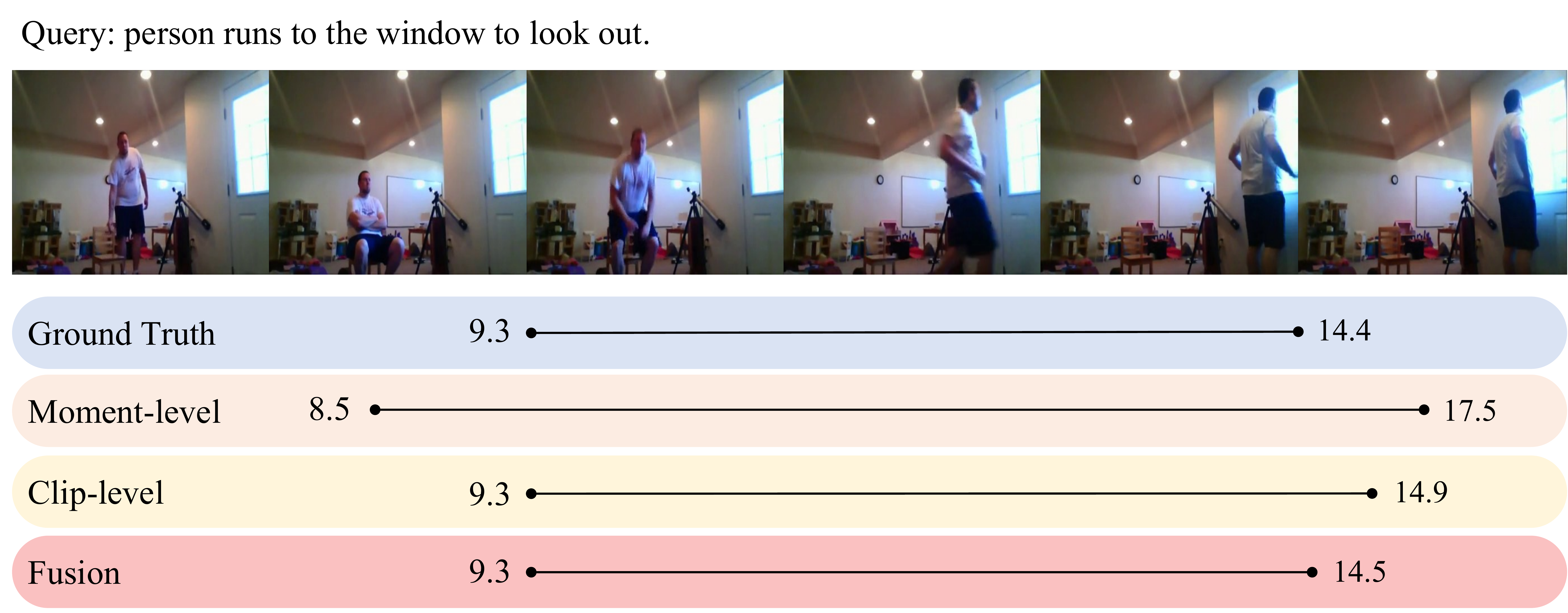}\label{v_i2}}
\end{center}
\caption{The prediction examples on Charades-STA with I3D features.}
\label{final}
\end{figure}

\bibliographystyle{ACM-Reference-Format}
\bibliography{sample-base}